\documentclass[11pt]{article}
\usepackage{amsmath, graphicx, enumerate, latexsym,amssymb}

\usepackage[dvips]{color}

\newtheorem{theorem}{Theorem}
\newtheorem{definition}{Definition}

\newtheorem{lemma}{Lemma}

\def\square{\hfill${\vcenter{\vbox{\hrule height.4pt \hbox{\vrule width.4pt
height7pt \kern7pt \vrule width.4pt} \hrule height.4pt}}}$}

\newenvironment{proof}{\noindent {\bf Proof.}\quad}{\square \vskip 12pt}

\begin{document}

\title{A Note on the Inapproximability of Correlation Clustering}

\author{
Jinsong Tan \thanks{Department of Computer \&
    Information Sciences, University of Pennsylvania, Philadelphia, PA
    19104. Email: {\tt jinsong@seas.upenn.edu }}
}
\date{}

\maketitle

\abstract{We consider inapproximability of the correlation
clustering problem defined as follows: Given a graph $G = (V,E)$
where each edge is labeled either "$+$" (similar) or "$-$"
(dissimilar), correlation clustering seeks to partition the
vertices into clusters so that the number of pairs correctly
(resp. incorrectly) classified with respect to the labels is
maximized (resp. minimized). The two complementary problems are
called \textsc{MaxAgree} and \textsc{MinDisagree}, respectively,
and have been studied on complete graphs, where every edge is
labeled, and general graphs, where some edge might not have been
labeled. Natural edge-weighted versions of both problems have been
studied as well. Let $\mathcal{S}$-\textsc{MaxAgree} denote the
weighted problem where all weights are taken from set
$\mathcal{S}$, we show that $\mathcal{S}$-\textsc{MaxAgree} with
weights bounded by $O(|V|^{1/2-\delta})$ essentially belongs to
the same hardness class in the following sense: if there is a
polynomial time algorithm that approximates
$\mathcal{S}$-\textsc{MaxAgree} within a factor of $\lambda =
O(\log{|V|})$ with high probability, then for any choice of
$\mathcal{S'}$, $\mathcal{S'}$-\textsc{MaxAgree} can be
approximated in polynomial time within a factor of $(\lambda +
\epsilon)$, where $\epsilon > 0$ can be arbitrarily small, with
high probability. A similar statement also holds for
$\mathcal{S}$-\textsc{MinDisagree}. This result implies it is hard
(assuming $\mathcal{NP\neq RP}$) to approximate unweighted
\textsc{MaxAgree} within a factor of $80/79-\epsilon$, improving
upon a previous known factor of $116/115-\epsilon$ by Charikar et.
al. \cite{Chari05}.\footnote{Throughout the paper, when we talk
about approximation factors we adopt the convention of assuming
the factor is greater than 1 for both maximization and
minimization problems.} \vspace{0.5em}

}

\section{Introduction}

\noindent Motivated by applications of document clustering,
Bansal, Blum and Chawla \cite{Ban04} introduced the correlation
clustering problem where for a corpus of documents, we represent
each document by a node, and an edge $(u,v)$ is labeled "+" or
"$-$" depending on whether the two documents are similar or
dissimilar, respectively. The goal of correlation clustering is
thus to find a partition of the nodes into clusters that agree as
much as possible with the edge labels. Specifically, there are two
complementary problems. \textsc{MaxAgree} aims to maximize the
number of agreements: the number of + edges inside clusters plus
the number of $-$ edges across clusters; on the other hand,
\textsc{MinDisagree} aims to minimize the number of disagreements:
the number of + edges across different clusters plus the number of
$-$ edges inside clusters. Correlation clustering is also viewed
as a kind of agnostic learning problem \cite{Kearns94} and seems
to have been first studied by Ben-Dor et al. \cite{Ben-Dor99} with
applications in computational biology; Shamir et al.
\cite{Shamir02} were the first to formalize it as a
graph-theoretic problem, which they called Cluster Editing. Since
Bansal et al.¡¯s independent introduction of this problem
\cite{Ban04}, it has been studied quite extensively in recent
years \cite{Ailon05,Chari05,Demaine03,Eman03,Giotis06,Swam04}.

\textsc{MaxAgree} and \textsc{MinDisagree} have been studied on
complete graphs, where every edge is labeled, and general graphs,
where some edge might not have been labeled. The latter captures
the case where a judge responsible for producing the labels is
unable to tell if certain pairs are similar or not. Also, it is
natural for the judge to give some `confidence level' for the
labels he produces; this leads to the natural edge-weighted
versions, which we call $\mathcal{S}$-\textsc{MaxAgree} and
$\mathcal{S}$-\textsc{MinDisagree} respectively, indicating the
edge weights are taken from set $\mathcal{S}$.

The various versions of correlation clustering are fairly well
studied. For complete unweighted case, Bansal {\it et al.}
\cite{Ban04} gave a PTAS for \textsc{MaxAgree} and Charikar {et
al.} \cite{Chari05} gave a 4-approximation for
\textsc{MinDisagree} and showed APX-hardness. For general weighted
graphs, an $O(\log{n})$-approximation algorithm was also given in
\cite{Chari05} for \textsc{MinDisagree}, and algorithms with the
same approximation factor were also obtained independently by
Demaine and Immorlica \cite{Demaine03}, and Emanuel and Fiat
\cite{Eman03}; a $\frac{1}{0.7664}$-approximation algorithm was
given for \textsc{MaxAgree} in \cite{Chari05}, and this was
improved by Swamy \cite{Swam04} with a
$\frac{1}{0.7666}$-approximation algorithm.

In this paper, we focus on the general graph case. Our main
contribution is to show $\mathcal{S}$-\textsc{MaxAgree} (resp.
$\mathcal{S}$-\textsc{MaxAgree}) with absolute values of weights
bounded by $O(|V|^{1/2-\delta})$ belongs to the same hardness
class in the following sense: if there is a polynomial time
algorithm that approximates $\mathcal{S}$-\textsc{MaxAgree} (resp.
$\mathcal{S}$-\textsc{MaxAgree}) within a factor of $\lambda =
O(\log{|V|})$ with high probability, then for any choice of
$\mathcal{S'}$, $\mathcal{S'}$-\textsc{MaxAgree} (resp.
$\mathcal{S}$-\textsc{MaxAgree}) can be approximated in polynomial
time within a factor of $(\lambda + \epsilon)$, for any constant
$\epsilon > 0$, with high probability. This result implies it is
hard (assuming $\mathcal{NP\neq RP}$) to approximate unweighted
\textsc{MaxAgree} within a factor of $80/79-\epsilon$, improving
upon a previous known factor of $116/115-\epsilon$ by Charikar,
Guruswami and Wirth \cite{Chari05}.

\begin{theorem}\label{Charikar05} {\bf (\cite{Chari05})} For every $\epsilon > 0$, it is
$\mathcal{NP}$-hard to approximate the weighted version of
\textsc{MaxAgree} within a factor of $80/79-\epsilon$.
Furthermore, it is $\mathcal{NP}$-hard to approximate the
unweighted version of \textsc{MaxAgree} within a factor of
$116/115-\epsilon$.
\end{theorem}

\section{Definitions and Notations}

We give definitions and notations in this section.

\begin{definition} {\textsc ($\mathcal{S}$-\textsc{MaxAgree)}} A
\textsc{MaxAgree} problem is called
$\mathcal{S}$-\textsc{MaxAgree} if all edge weights are taken from
set $\mathcal{S}$. An element in $\mathcal{S}$ can be either a
constant or some function in the size of the input graph.
\end{definition}

\noindent $\mathcal{S}$-\textsc{MinDisagree} is defined likewise.
We assume 0 is always an element in $\mathcal{S}$ as we are
interested in the problem on general graphs in this paper.
Assigning weight 0 to non-edges allows us to view any general
graph as a complete one.

\begin{definition}\label{Roll} {(N-fold Roll)} Given a graph $G = (V,E)$ where
$V = \{v_1, v_2, ..., v_n\}$. Let $(N-1)$ be a multiple of
$(n-1)$, an $N$-fold roll (denoted by $G^N$) of $G$ is created by
embedding multiple copies of $G$ into an $N$ by $n$ grid where
there are $N$ parallel copies of $V$ and a node $v_{ij}$
corresponds to $v_j$ in the $i$th copy of $V$.

Edges of $G^N$ are created as follows. For any pair of nodes
$v_{i_1j_1}$ and $v_{i_2j_2}$, where $i_1, i_2 \in \{1,2,...,N\}$,
$j_1,j_2 \in \{1,2,...,n\}$. Define the `wrapped-around' vertical
distance of the two nodes

$$
d(v_{i_1j_1},v_{i_2j_2}) =
\begin{cases}
(i_2-i_1 \bmod{N}) \qquad & (j_1 \leq j_2) \\
\infty & (\mbox{otherwise})
\end{cases}
$$ A pair $(v_{i_1j_1}, v_{i_2j_2})$ is called a grid-bone if and only if $$
\begin{array}{lllr}
& 1) & j_1 \neq j_2; \mbox{ and } \\
& 2) & \frac{d(v_{i_1j_1},v_{i_2j_2})}{j_2-j_1} \in \{0,1,...,
\frac{N-1}{n-1} \}.
\end{array}
$$ A grid-bone $(v_{i_1j_1}, v_{i_2j_2})$ is an edge identical to
$(v_{j_1}, v_{j_2})$ (resp. non-edge), depending on whether
$(v_{j_1}, v_{j_2})$ is an edge (resp. non-edge) in $G$. All
non-grid-bone pairs $(v_{i_1j_1}, v_{i_2j_2})$ are non-edges.
\end{definition}

\noindent Note by construction $G^N$ consists of exactly
$N(\frac{N-1}{n-1}+1) > \frac{N^2}{n}$ duplicates of $G$. It is
conceptually easier to see this by indexing each duplicate with
pair $(i, c)$, where $i \in \{1, 2, ..., N\}$ indexes the $N$
parallel copies of $V$ and $c \in \{0, 1, ..., \frac{N-1}{n-1}\}$
can be thought of as the `slope' of the grid-bones in this copy.
More precisely, duplicate $(i, c)$ consists of nodes
$$\{v_{(i \bmod N)1}, v_{(i+c \bmod N)2}, v_{(i+2c \bmod N)3},
..., v_{(i+(n-1)c \bmod N)n}\}$$ For our purpose that will be
evident in the rest of the paper and for the sake of simpler
analysis, we assume w.l.o.g. that there are exactly
$\frac{N^2}{n}$ duplicates of $G$. Note this can be thought of as
erasing all edges on (any) excessive $N(\frac{N-1}{n-1}+1) -
\frac{N^2}{n}$ duplicates.

In this construction, we obtain $\frac{N^2}{n}$ disjoint
duplicates of $E$ from just $N$ disjoint duplicates of $V$, this
asymptotic gap is crucial in our proof of the main technical
results (i.e. Lemma \ref{maxAgree} and \ref{minDisagree}). We will
discuss why we need this gap in the proof of Lemma \ref{maxAgree}.

\begin{definition} {$( \mathcal{S}$-to-$\{-\alpha,0,\beta \}$ randomized rounding)}
\vspace{0.2em}

\noindent{\bf Input:} An instance of
$\mathcal{S}$-\textsc{MaxAgree}
($\mathcal{S}$-\textsc{MinDisagree}) on general graph $G = (V,
E)$, where w.l.o.g. it is assumed that $\gamma \leq 1$, $\forall~
\gamma \in \mathcal{S}$; and $\alpha, \beta \geq 1$.
\vspace{0.2em}

\noindent{\bf Output:} The same graph with the following
randomized rounding. For each edge of weight $\gamma > 0$ (resp.
$\gamma < 0$), round $\gamma$ to either 0 or $\beta$ (resp. either
$-\alpha$ or 0) independently and identically at random with
expectation being $\gamma$.
\end{definition}

Denote by $w(\cdot)$ the weight function before rounding, and
$w'(\cdot)$ the one after rounding. We slightly abuse notation
here by allowing both weight functions to take edges and
clusterings as parameter. For a clustering $C$, denote by
$w'_{\gamma}(C)$ the total post-rounding weight of $C$ contributed
by former-$\gamma$-edges.

\begin{definition} {(Contributing)} Given an
$\mathcal{S}$-\textsc{MaxAgree} (resp.
$\mathcal{S}$-\textsc{MinDisagree}) instance and a clustering $C$,
we call an edge $(i,j)$ of weight $\gamma$ a {\it contributing}
edge iff $\gamma > 0$ (resp. $\gamma < 0$) and $(i,j)$ is inside a
cluster of $C$, or $\gamma < 0$ (resp. $\gamma > 0$) and $(i,j)$
is cross different clusters of $C$.
\end{definition}

\section{Main Theorems}

Given an $\mathcal{S}$-{\sc MaxAgree} (resp. $\mathcal{S}$-{\sc
MinDisagree}) instance, first construct an $N$-fold roll $G^N =
(V^N, E^N)$, and then apply the
$\mathcal{S}$-to-$\{-\alpha,0,\beta \}$ randomized rounding on
$G^N$. If we solve the $\{-\alpha,0,\beta \}$ instance on $G^N$,
the solution clustering $C$ implies a total of $\frac{N^2}{n}$
(not necessarily distinct) ways to cluster $G$, one for each of
the $\frac{N^2}{n}$ duplicates of $G$. To see this, note $C$ is
simply a partition of $V^N$, and this partition induces a
partition, thus a clustering, on each of the $\frac{N^2}{n}$
duplicates of $G$. We call each of these clusterings a {\em
candidate solution} to the initial $\mathcal{S}$-{\sc MaxAgree}
(resp. $\mathcal{S}$-{\sc MinDisagree}) instance on $G$ and denote
them as $C_1, C_2, ..., C_{\frac{N^2}{n}}$.

Note although these $\frac{N^2}{n}$ duplicates of $G$ share nodes
of $G^N$, their edge sets are disjoint. In fact, these
$\frac{N^2}{n}$ duplicates of $E$ form a partition of $E^N$. Lemma
\ref{lemma:crucial} is immediate.

\begin{lemma}\label{lemma:crucial} For both $\mathcal{S}$-{\sc MaxAgree} and $\mathcal{S}$-{\sc
MinDisagree}, $w(C) = \sum_{i=1}^{N^2/n} w(C_i)$.
\end{lemma}

Our next lemma says that if an edge is not contributing before
rounding, it must not be contributing after rounding. Therefore,
to calculate the weight of $C$ both before and after the rounding,
we only need to concern ourself with the same set of edges.

\begin{lemma}\label{lemma:random} For both $\mathcal{S}$-{\sc MaxAgree} and $\mathcal{S}$-{\sc
MinDisagree}, let $E(C)$ be the set of contributing edges of $C$
before randomized rounding is applied to $G^N$, i.e. $w(C) =
\sum_{e\in E(C)} w(e)$. Then after rounding, the new weight of $C$
is still a summation over the same set of edges, i.e. $w'(C) =
\sum_{e\in E(C)} w'(e)$.
\end{lemma}
\begin{proof} This follows from the observation that positive
edges are rounded to have either positive or zero weights, and
negative edges are rounded to have either negative or zero
weights.
\end{proof}

We are now ready to give our main technical result in Lemma
\ref{maxAgree}. We concern ourself only with
$\mathcal{S}$-\textsc{MaxAgree} here; a similar result holds for
$\mathcal{S}$-\textsc{MinDisagree} and is given in Lemma
\ref{minDisagree}.

\begin{lemma} \label{maxAgree} Given an
$\mathcal{S}$-\textsc{MaxAgree} instance $G = (V, E)$, let $G^N =
(V^N, E^N)$ be the $N$-fold roll of $G$ with
$\mathcal{S}$-to-$\{-\alpha, 0, \beta\} $ randomized rounding
applied. If \vspace{0.5em}

1. $\alpha + \beta = O((Nn)^{1/2-\delta})$, where $\delta \in (0,
\frac{1}{2}]$; and

2. there is a $\lambda$-approximation algorithm for
$\{-\alpha,0,\beta\}$-\textsc{MaxAgree}, where $\lambda =
O(\log{n})$ \vspace{0.5em}

\noindent then for any arbitrarily small number $\epsilon > 0$
there exists a polynomial time algorithm that approximates
$\mathcal{S}$-\textsc{MaxAgree} within a factor of
$(\lambda+\epsilon)$ with probability at least $\frac{1}{2}$.
\end{lemma}

\begin{proof} For any $\gamma \in \mathcal{S}$, let $X^{(\gamma)}$
denote the random variable representing the new weight of a
former-$\gamma$-edge after rounding. Define random variable
$Y^{(\gamma)} = X^{(\gamma)} -\gamma$; clearly $E[Y^{(\gamma)}] =
0$. Note it is assumed w.l.o.g. that $|\gamma| \leq 1$, $\forall~
\gamma \in \mathcal{S}$.

Suppose there is a polynomial time algorithm $\mathcal{A}$ that
approximates $\{-\alpha,0,\beta\}$-\textsc{MaxAgree} within a
factor of $\lambda$, we can then run $\mathcal{A}$ on $G^N$, the
output clustering $C_2^*$ corresponds to $\frac{N^2}{n}$ ways to
cluster $G$ (not necessarily all distinct). Let $C_1^*$ be the
most weighted among these $\frac{N^2}{n}$ clusterings of $G$, in
the rest of the proof we show that with high probability, $C_1^*$
is a $(\lambda+\epsilon)$-approximation of
$\mathcal{S}$-\textsc{MaxAgree} on $G$ for any fixed $\epsilon$.

Denote by $\mathbb{E}$ the bad event that $C_2^*$ does not imply a
$(\lambda+\epsilon)$-approximation on $G$, i.e. $C_1^*$ is not a
$(\lambda+\epsilon)$-approximation. Let $C'$ be an arbitrary
clustering of $G^N$ that does not imply a
$(\lambda+\epsilon)$-approximation on $G$. Denote by
$\mathbb{E}(C')$ the event that $C'$ becomes a
$\lambda$-approximation on $G^N$ after rounding. Since there are
at most $(Nn)^{Nn}$ distinct clusterings of $G^N$, by union bound
we have $ Pr\left\{ \mathbb{E} \right\} \leq e^{Nn \ln{Nn}} \cdot
Pr\left\{ \mathbb{E}(C') \right\}$. (We note that the randomness
of event $\mathbb{E}(C')$ comes from the randomized rounding and
the randomness of event $\mathbb{E}$ comes from both the
randomized rounding and the internal randomness of $\mathcal{A}$.)

Let the weight of an optimal clustering $U$ of $G$ be $K$, denote
by $U^N$ the corresponding duplication clustering in $G^N$. That
is, $U^N$ has the same number of clusters as $U$, and there is a
one-to-one mapping between the two sets of clusters such that a
node $v_j$ is in a cluster of $U$ if and only if all its $N$
duplicates, $v_{1j}, v_{2j}, ..., v_{Nj}$, are in the
corresponding cluster of $U^N$. We now proceeds to prove that
event $\mathbb{E}(C')$ happens with negligible probability. Before
delving into the details, we first offer a high level discussion
of the idea behind the proof. \vspace{0.5em}

\noindent{\bf Intuition Behind the Proof. } {\em Since $U$ is an
optimal clustering of $G$, by Lemma \ref{lemma:crucial} it is easy
to see that $U^N$ is an optimal clustering of $G^N$ before
randomized rounding and its weight is $\frac{KN^2}{n}$. $C'$ is an
arbitrary but fixed clustering. Since it does not imply a
$(\lambda+\epsilon)$-approximation on $G$, it must be the case
that before rounding the weight of $C'$ on $G^N$ is less than
$\frac{KN^2}{(\lambda+\epsilon)n}$. Since $\epsilon$ is a fixed
constant, this leaves a gap between
$\frac{KN^2}{(\lambda+\epsilon)n}$ and $\frac{KN^2}{\lambda n}$.
By Lemma \ref{lemma:random} the expectation of the new weight of
$U^N$ is $\frac{KN^2}{n}$ and that of $C'$ is at most
$\frac{KN^2}{(\lambda+\epsilon)n}$. Therefore for the bad event
$\mathbb{E}(C')$ to happen either $C'$ has to be really lucky in
the rounding so that its new weight ends up hitting as high as
$\frac{KN^2}{\lambda n}$, or $U^N$ has to be really unlucky in the
rounding so that its new weight ends up touching as low as
$\frac{\lambda KN^2}{(\lambda + \epsilon)n}$, or mostly likely
some sort of combination of the two. Whichever case happens, the
common thing shared is that one has to rely on pure chance to
close the gap. And we show that by setting $N = poly(n)$
sufficiently large, this happens with negligible probability. In
fact, the probability of $\mathbb{E}(C')$ is so small that even
$(Nn)^{Nn}$ times of it is still negligible.  } \vspace{0.5em}

We now resume the proof. For any $\gamma \in \mathcal{S}$, and a
clustering $C$ of $G^N$, denote by $E(C, \gamma)$ the set of
former-$\gamma$-edges that are contributing in $C$ {\em before}
rounding. If $\mathbb{E}(C')$ happens, then $$
\begin{array}{cc}
\displaystyle \sum_{\gamma \in \mathcal{S}} w'_{\gamma}(C') \geq \dfrac{1}{\lambda} \left( \sum_{\gamma \in \mathcal{S}} w'_{\gamma}(U^N) \right) \\
\displaystyle \sum_{\gamma \in \mathcal{S}} |E(C', \gamma)| \cdot |\gamma| < \dfrac{K N^2 / n}{\lambda+\epsilon} = \dfrac{1}{\lambda+\epsilon} \left( \sum_{\gamma \in \mathcal{S} } { |E(U^N, \gamma)| \cdot |\gamma| } \right) \\
\end{array}$$ where the first inequality follows because $C'$ is a
$\lambda$-approximation of $G^N$, and $\sum_{\gamma \in
\mathcal{S}} w'_{\gamma}(\cdot)$ is the total weight of a
clustering after rounding; the second inequality follows from
Lemma \ref{lemma:crucial} and the fact that each of the
$\frac{N^2}{n}$ candidate solutions implied by $C'$ has weight
less than $\frac{K}{\lambda+\epsilon}$. Simple manipulation of the
two inequalities above yields $$ S_1 - \dfrac{S_2}{\lambda}
> \dfrac{\epsilon}{\lambda (\lambda+\epsilon)} \left(
\displaystyle \sum_{\gamma \in \mathcal{S}}{|E(U^N, \gamma)| \cdot
|\gamma|} \right) = \dfrac{\epsilon K N^2/n}{\lambda
(\lambda+\epsilon)}
$$ where $S_1 = \left( \displaystyle \sum_{\gamma \in \mathcal{S}} {\left(
w'_{\gamma}(C') - |E(C', \gamma)| \cdot |\gamma| \right)} \right)$
and $S_2 = \dfrac{1}{\lambda} \left( \displaystyle \sum_{\gamma
\in \mathcal{S} }{ (w'_{\gamma}(U^N) - |E(U^N, \gamma) | \cdot
|\gamma| ) } \right)$. Since $\lambda = O(\log{(Nn)}) =
O(\log{n})$ and $K \geq 1$, when $n$ is sufficiently large, $S_1 -
\dfrac{S_2}{\lambda} \geq \dfrac{\epsilon N^2}{n^2}$. This implies
$$
Pr\{ \mathbb{E}(C') \} \leq  Pr \left\{ S_1 - \dfrac{S_2}{\lambda}
> \dfrac{\epsilon N^2}{n^2} \right\}
$$ Note the expectation of both $S_1$ and $S_2$ are 0, therefore
so is the linear combination $S_1 - S_2/\lambda$; in the following
we argue that the probability for $S_1 - S_2/\lambda$ to deviate
from its mean by $\epsilon N^2 / n^2$ is negligible when $N$ is
sufficiently large.

For any $\gamma \in \mathcal{S}$, let $z_1(\gamma) = | E(C',
\gamma) - E(U^N, \gamma) |$ be the number of former-$\gamma$-edges
contributing in $C'$ but not $U^N$ before rounding. Similarly,
define $z_2(\gamma) = | E(U^N) - E(C') |$ and $z_3(\gamma) =
|E(U^N) \cap E(C') |$. We have $$
\begin{array}{llr}
& Pr \left\{ S_1 - \dfrac{S_2}{\lambda}

> \dfrac{\epsilon N^2}{n^2} \right\} \\

= & Pr \left\{\displaystyle \sum_{\gamma \in \mathcal{S}} {  \left
( \sum_{i=1}^{z_1(\gamma)}{Y_i^{(\gamma)}} + \dfrac{1}{\lambda}
\sum_{j=1}^{z_2(\gamma)}{(-Y_j^{(\gamma)})} +
\dfrac{\lambda-1}{\lambda}
\sum_{k=1}^{z_3(\gamma)}{Y_k^{(\gamma)}}   \right) }
> \dfrac{\epsilon N^2}{n^2} \right\} \\

\leq & Pr \left\{\displaystyle \sum_{\gamma \in \mathcal{S}} {

 \left ( \sum_{i=1}^{z_1(\gamma)}{Y_i^{(\gamma)}} +
\sum_{j=1}^{z_2(\gamma)}{(-Y_j^{(\gamma)})} +
\sum_{k=1}^{z_3(\gamma)}{Y_k^{(\gamma)}}   \right) }
> \dfrac{\epsilon N^2}{n^2} \right\} & \mbox{($\lambda > 1$)} \\

\leq & \displaystyle \sum_{\gamma \in \mathcal{S}} { \sum_{h\in
\{1,2,3\}} {

\left( Pr \left\{ \displaystyle
\sum_{i=1}^{z_h(\gamma)}{(-1)^{(h-1)}Y_i^{(\gamma)}} >
\dfrac{\epsilon N^2}{3n^2 |\mathcal{S}|} \right\}

\right) } } & \mbox{(union bound)} \\

\leq & \displaystyle \sum_{\gamma \in \mathcal{S}} { \sum_{h \in
\{1,2,3\}}^{} { \left( \exp \left( -2z_h(\gamma)
\left(\dfrac{\epsilon N^2}{3n^2 |\mathcal{S}| \cdot z_h(\gamma) \cdot (\alpha+\beta)} \right)^2 \right) \right) } } & \mbox{(Hoeffding bound)} \\

\leq & 3 |\mathcal{S}| \cdot \exp \left( -c_1 \cdot
\dfrac{N^2/n}{n^8 (\alpha+\beta)^2}  \right) & \mbox{($|\mathcal{S}|\leq n^2$, $z_h(\gamma) \leq N^2 n$)}\\

\end{array}
$$ where $c_1$ is some constant. Since we allow $\alpha+\beta = O((Nn)^{(1/2-\delta)})$ and want
$(Nn)^{Nn} \cdot Pr(\mathbb{E}(C'))$ to be negaligible, it is now
clear why we need $N^2/n$ duplicates of $E$ and thus the {\it
N-fold roll} construction given in Definition \ref{Roll}. In
contrast, had we adopted a naive construction with $N$ isolated
duplicates of $G$, there will be only $N$ duplicates of $E$; and
it is readily verified that this is insufficient to prove that
$(Nn)^{Nn} \cdot Pr(\mathbb{E}(C'))$ is negligible.

Now set $N = n^{6/\delta}$, we have
$$ Pr\left\{\mathbb{E} \right\} \leq (Nn)^{Nn} \cdot Pr \left\{
\mathbb{E}(C') \right\} \leq 3 n^2 \cdot \exp{\left( (6/\delta +1)
n^{6/\delta+1} \ln{n} - c_2 \cdot n^{(6/\delta +2 + 2\delta)}
\right)}
$$ for some constant $c_2$. Note this probability is bounded by $\frac{1}{2}$
as the input size $n$ is sufficiently large. Therefore we have
obtained a polynomial time algorithm that approximates
$\mathcal{S}$-\textsc{MaxAgree} within a factor of
$\lambda+\epsilon$ with probability at least $\frac{1}{2}$.
\end{proof}

\noindent We give a similar result for
$\mathcal{S}$-\textsc{MinDisagree} in Lemma \ref{minDisagree}, the
proof follows essentially exactly the same construction and
analysis as Lemma \ref{maxAgree} so we only give a high level
discussion without duplicating the proof.

\begin{lemma} \label{minDisagree} Given an
$\mathcal{S}$-\textsc{MinDisagree} instance $G = (V, E)$, let $G^N
= (V^N, E^N)$ be the $N$-fold roll of $G$ with
$\mathcal{S}$-to-$\{-\alpha, 0, \beta\} $ randomized rounding. If
\vspace{0.5em}

1. $\alpha + \beta = O((Nn)^{1/2-\delta})$, where $\delta \in (0,
\frac{1}{2}]$; and

2. there is a $\lambda$-approximation algorithm for
$\{-\alpha,0,\beta\}$-\textsc{MinDisagree}, where $\lambda =
O(\log{n})$ \vspace{0.5em}

\noindent then for any arbitrarily small number $\epsilon > 0$
there exists a polynomial time algorithm that approximates
$\mathcal{S}$-\textsc{MinDisagree} within a factor of
$(\lambda+\epsilon)$ with probability at least $\frac{1}{2}$.
\end{lemma}
\begin{proof} (Sketch) We define $U^N$ and $C'$ analogously as in
that of Lemma \ref{maxAgree}. The weight of $U^N$ before rounding
is $\frac{KN^2}{n}$, and the weight of $C'$ before rounding is
greater than $\frac{(\lambda+\epsilon) KN^2}{n}$. Again since
$\epsilon$ is a fixed constant, there is a gap between
$\frac{(\lambda+\epsilon) KN^2}{n}$ and $\frac{\lambda KN^2}{n}$.
For $C'$ to be a $\lambda$-approximation after rounding, its new
weight must necessarily be at most $\lambda$ times of the new
weight of $U^N$. Since the expectation of the new weight of $U^N$
is $\frac{KN^2}{n}$ and that of $C'$ is greater than
$\frac{(\lambda+\epsilon) KN^2}{n}$, again we need to rely on
chance to close this gap of $\frac{\epsilon KN^2}{n}$. By applying
a similar analysis as in Lemma \ref{maxAgree} we can show that
even $(Nn)^{Nn}$ times of this probability, which upper bounds the
probability of the bad event that a $\lambda$-approximation on
$G^N$ does not imply a $(\lambda+\epsilon)$-approximation on $G$,
is negligible.
\end{proof}

\noindent Lemma \ref{maxAgree} and \ref{minDisagree} leads to the
following theorem.

\begin{theorem}\label{mainThm} If $\mathcal{S}$-\textsc{MaxAgree} (resp. $\mathcal{S}$-\textsc{MinDisagree}) is $\mathcal{NP}$-hard to
approximate within a factor of $\lambda$ ($\lambda = O(\log{n})$)
for any specific choice of $\mathcal{S}$, then for any choice of
$\mathcal{S'}$, where $\max_{\gamma \in \mathcal{S'}}|\gamma| =
O(n^{1/2-\delta})$ for some $\delta \in (0,\frac{1}{2}]$, no
polynomial time algorithm, possibly randomized, can approximate
$\mathcal{S'}$-\textsc{MaxAgree} (resp.
$\mathcal{S'}$-\textsc{MinDisagree}) within a factor of $\lambda +
\epsilon$ with probability at least $\frac{1}{2}$ unless
$\mathcal{NP=RP}$.
\end{theorem}
\begin{proof} This follows from Lemma \ref{maxAgree} and
\ref{minDisagree} by setting $\alpha = - \min{ \mathcal{S} }$ and
$\beta = \max{\mathcal{S} }$.
\end{proof}

\noindent In particular, invoking the result by Charikar {\it et
al.} in Theorem \ref{Charikar05} leads to the following improved
inapproximability result.

\begin{theorem}
No polynomial time algorithm, possibly randomized, can approximate
unweighted version of \textsc{MaxAgree} in general graphs within a
factor of $80/79-\epsilon$ unless $\mathcal{NP=RP}$.
\end{theorem} \vspace{0.5em}

\noindent{\bf Acknowledgeement.} The author would like to thank
Tanmoy Chakraborty and the anonymous reviewers for their valuable
comments and suggestions that helped to improve the presentation
of the paper.

\end{document}